\documentclass[11pt]{article}

\usepackage{amsmath,amsthm,amsfonts}

\usepackage{amssymb}

\usepackage{epsfig}

\usepackage{rotating}

\usepackage{subfigure}

\usepackage{float}

\usepackage{algorithm}

\usepackage{algpseudocode}

\usepackage{authblk}
\usepackage{diagbox}
\usepackage{fixltx2e}

\begin{document}

\title{Enhancing classification accuracy through chaos}

\author[]{Panos Stinis}

\affil[]{Advanced Computing, Mathematics and Data Division, Pacific Northwest National Laboratory, Richland WA 99354}

\renewcommand\Affilfont{\itshape\small}

\date {}

\maketitle

\begin{abstract}
We propose a novel approach which exploits chaos to enhance classification accuracy. Specifically, the available data that need to be classified are treated as vectors that are first lifted into a higher-dimensional space and then used as initial conditions for the evolution of a chaotic dynamical system for a prescribed temporal interval. The evolved state of the dynamical system is then fed to a trainable softmax classifier which outputs the probabilities of the various classes. As proof-of-concept, we use samples of randomly perturbed orthogonal vectors of moderate dimension (2 to 20), with a corresponding number of classes equal to the vector dimension, and show how our approach can reduce the cross-entropy loss by several orders of magnitude and the number of training epochs by at least an order of magnitude, compared to a standard softmax classifier which operates on the original vectors, as well as a softmax classifier which only lifts the vectors to a higher-dimensional space without evolving them. We also provide an explanation for the improved performance of the chaos-enhanced classifier and a selection process for the optimal chaotic evolution interval.   
\end{abstract}

\section*{Introduction}
Classification is one of the two fundamental supervised machine learning (ML) tasks \cite{kotsiantis2006machine,lecun2015deep}, the other being regression \cite{lathuiliere2019comprehensive,karniadakis2021physics}. It is a vast research topic with applications ranging from health \cite{an2023comprehensive} to finance \cite{gerlein2016evaluating} to science \cite{tarca2007machine,carleo2019machine,choudhary2022recent}. The objective of ML-based classification is to train a ML tool to assign the correct class (label) to each member of a dataset. Usually, ML-based classifiers assign, for each member of a dataset, a probability for the member to belong to each of the available classes. 

Due to the pervasive nature of classification, there are many algorithms that have been developed for this task and an exhaustive comparison is beyond the scope of the current work. Instead, we will reduce the ML-classification problem to its bare essentials and offer a novel approach which can both accelerate training and improve the training/testing accuracy. The approach we will discuss can be applied directly to the raw input data as a way of promoting separation of the datapoints from different classes so that they are classified more accurately. It is analogous to the separability of data promoted by neural networks through the extraction of hierarchical features as information is processed through the hidden layers \cite{he2023law}. We note that the approach can also be applied to the latent space data representation provided by an existing ML-based algorithm (before the final linear output layer which connects to a softmax classifier that assigns probabilities to the various classes), although this will not be pursued here. 

We consider a dataset consisting of randomly perturbed orthogonal vectors of a prescribed dimension. The correct class for each of the perturbed orthogonal vectors is the direction corresponding to the unperturbed vector. The baseline model we will use consists of a trainable softmax classifier to which we feed the randomly perturbed orthogonal vectors. The novel approach we propose consists of two stages: (i) lifting the vector to a higher dimension, and (ii) identifying the lifted vector as the initial state of a chaotic dynamical system and evolving it for a prescribed temporal interval before it is processed by the trainable softmax classifier. Our purpose in this work is to disentangle the effect that the two stages can have on the classification training time and accuracy, so that the observed improvement over the baseline model, which can be significant, can be explained. 

Section \ref{chaos} presents the various constructions including the baseline classifier model, the lifting to higher dimensions and the chaos-enhanced model. Section \ref{results} contains numerical results for the various constructions along with an explanation of the improved performance of the chaos-enhanced classifier and a selection process for the optimal chaotic evolution interval, while Section \ref{discussion} offers a discussion of the proposed approach and directions for future work.
 
\section{Chaos-enhanced classification}\label{chaos}
Consider a collection $V_N=[v_1,v_2,\ldots,v_N]$ of $N$ vectors of dimension $m,$ each being a random perturbation of a canonical basis vector in $\mathbb{R}^m.$ The canonical basis in $m$ dimensions consists of the vectors $e_i = [\underbrace{0,\ldots,0}_{i-1},1,\underbrace{0,\ldots,0}_{m-i}]^T,$ for $i=1,\ldots,m,$ with all $0$s except for the 1 at the $i$-th coordinate. For symmetry, we will assume that the $m$ canonical basis vectors are equally represented in the collection, say with $N_{class}$ vectors each, so $N=m \times N_{class}.$ Each of the $N$ vectors is formed by adding to to all the coordinates of the orthogonal vectors samples from independent, normally distributed random variables with mean zero and prescribed variance. Thus, each of the $N$ vectors in $V_N$ can be written as $v_{nk} = e_{i_nk} + \epsilon_{nk},$ for $n=1,\ldots,N$ and $k=1,\ldots,m,$ where $\epsilon_{nk} \sim \mathcal{N} (0,\sigma^2)$ with $\sigma^2$ prescribed and $i_n,$ with $i_n$ taking one value from $1,\dots,m,$ is the index corresponding to the canonical vector that we use to build the $n$-th sample in the collection.

{\bf Baseline model.} The baseline model consists of a trainable softmax classifier to which each of the $N$ vectors are fed. Since we consider the $N$ vectors coming from random perturbations of the canonical basis vectors in $\mathbb{R}^m,$ we consider $m$ classes, one for each of the $m$ orthogonal directions in $\mathbb{R}^m.$ As a result, the trainable weights of the linear output layer of the softmax classifier are the elements of a $m \times m$ matrix $W,$ denoted by $w_{ij}$ for $i,j =1,\ldots,m.$ For a vector $v_n$ from the collection of $N$ vectors, the softmax classifier takes as input the logits $z_{ni} = \sum_{j=1}^m W_{ij}v_{nj}$ for $i=1,\ldots,m,$ and outputs the $m$ probabilities 
\begin{equation}\label{softmax_baseline}
p_{ni} = \frac{e^{z_{ni}}}{\sum_{i=1}^m e^{z_{ni}}}, \;\; \text{for} \;\; i=1,\ldots,m,
\end{equation}
that the vector $v_n$ belongs to each of the $m$ classes. We employ cross-entropy as the loss function to train the elements of $W$ and use the Adam optimizer for training. The cross-entropy loss for the $N$ vectors is defined as 
\begin{equation}\label{loss_baseline}
L_N = - \frac{1}{N} \sum_{n=1}^N \sum_{l=1}^m \chi_{nl} \log p_{nl},
\end{equation}
where $\chi_{nl}=1$ if $l=i_n$ and 0 otherwise, where $i_n$ is the true class of the $n$-th sample, and $p_{nl}$ is provided by \eqref{softmax_baseline}. The derivative of the cross-entropy loss function with respect to an element of the weight matrix $W$ is given by 
\begin{equation}\label{loss_gradient_baseline}
\frac{\partial L_N}{\partial w_{ij}} = - \frac{1}{N} \sum_{n=1}^N [\chi_{ni} - p_{ni}] v_{nj}, \;\; \text{for} \;\; i,j=1,\ldots,m.
\end{equation}
Since the true classes of the vectors in $V_N$ are along the directions of the canonical basis vectors in $\mathbb{R}^m,$ the training of the baseline model aims to suppress the random perturbations along the $m$ directions (as can be seen by \eqref{loss_gradient_baseline}).

\begin{algorithm}[H] 
	\caption{Baseline classification algorithm}
	\begin{algorithmic}[1]
		\Require Number of classes $m,$ data $V_N$ (vectors of dimension $m$) where  $N = m \times N_{class}$ ($N_{class}$ orthogonal vectors from each of the $m$ classes perturbed with $\mathcal{N}(0,\sigma^2)$ noise, $\sigma^2$ is user-prescribed), number of training data $N_{train}$ and test data $N_{test}$ with $N_{train} + N_{test} = N,$ class labels $\chi$ for the vectors in $V_N,$ mini-batch size $N_b,$ initial learning rate for Adam optimizer $\eta,$ number of training epochs $N_{epochs}$
		\State \textbf{Training:}
		\State Initialization of the $m \times m$ weight matrix $W$
		\For{each training epoch $ n = 1, 2, \dots, N_{epochs} $}
		\State Pick a mini-batch from the training data $V_{N_{train}}$
		\State Assign class probabilities to the mini-batch using a softmax classifier
		\State Use the Adam optimizer with cross-entropy loss function to update the weight matrix $W$
		\State Repeat until the epoch is completed
        		\EndFor
	\end{algorithmic}
	\label{algorithm_baseline}
\end{algorithm}

{\bf Lifting-enhanced model.} As we have already discussed, the proposed approach can be divided into two stages: (i) lifiting each of the $N$ vectors to a higher dimension and (ii) evolving the lifted vector using a chaotic dynamical system. The first stage can be considered in isolation from the second one, since it can be beneficial for classification to lift, within reason, the dimensionality of the vectors to be classified \cite{bickel2004some,fan2008high,ghaddar2018high}. There are various methods to lift a vector to higher dimensions, e.g., zero-padding, polynomial feature mapping, tensor product. Each of the $N$ vectors of dimension $m$ is lifted to a vector of dimension $m_{lift}.$ As a result of the lifting process, the linear output layer weight matrix $W^{lift}$ has dimensions $m \times m_{lift}.$ For the $n$-th original (unlifted) vector $v_n=[v_{n1},\ldots,v_{nm}]^T,$ the corresponding lifted vector, denoted by $v_n^{lift},$ is defined as
\begin{equation}\label{lift}
v_n^{lift} = [\eta_1,\eta_2,\eta_3, v_{n1},\ldots,v_{nm}, \eta_4,\ldots,\eta_{m_{lift}-m}]^T,
\end{equation}
where $\eta_1,\ldots,\eta_{m_{lift}-m}$ are random linear combinations of the $m$ coordinates. Specifically, $\eta_j = \sum_{l=1}^m \epsilon_l v_{nl}$ where $ \epsilon_l \sim \mathcal{N}(0,\frac{1}{m_{lift}}).$ The specific choice of the lifting construction is due to the structure of the Lorenz 96 model that is used to evolve the lifted vector, namely the need to enforce periodicity (see \eqref{lorenz2} below). If the constraints from the Lorenz 96 system were not there, we could have picked a simpler lifting construction e.g., 
\begin{equation}\label{lift_simple}
v_n^{lift} = [v_{n1},\ldots,v_{nm}, \eta_1,\ldots,\eta_{m_{lift}-m}]^T.
\end{equation}
We note that the chosen lifting approach is not optimized, but does imbue the added coordinates with randomly distorted information from the $m$ coordinates of the original vector. Also, the lifting of each vector to one which preserves the $m$ coordinates of the original vector but adds (small) random perturbations in the remaining coordinates can be understood as a random regularization mechanism. This is because the weight matrix $W^{lift}$ will be trained to map from $\mathbb{R}^{m_{lift}}$ to the $m$ classes, thus it will be trained to suppress random perturbations in the $m_{lift}-m$ dimensions (in addition to learning to suppress random perturbations in the $m$ coordinates like the baseline model). Improving training accuracy through adding random perturbations to data is a well-known practice for neural networks \cite{nishi2021augmentation,an2025exploring,jacobsen2025staying,stinis2020enforcing}. When we present the numerical results (Section \ref{results}), we provide a more quantitative explanation as to how the lifting to a higher dimensional space aids the classification task.

Due to the lifting of the vectors to $m_{lift}$ dimensions, the logits $z_{ni}^{lift} = \sum_{j=1}^{m_{lift}} W_{ij}^{lift}v_{nj}^{lift}$ for $i=1,\ldots,m,$ and softmax outputs the $m$ probabilities 
\begin{equation}\label{softmax_lift}
p_{ni}^{lift} = \frac{e^{z_{ni}^{lift}}}{\sum_{i=1}^m e^{z_{ni}^{lift}}}, \;\; \text{for} \;\; i=1,\ldots,m,
\end{equation}
that the vector $v_n^{lift}$ belongs to each of the $m$ classes. The loss function is given by 
\begin{equation}\label{loss_lift}
L_N^{lift} = - \frac{1}{N} \sum_{n=1}^N \sum_{l=1}^m \chi_{nl}^{lift} \log p_{nl}^{lift},
\end{equation}
where $\chi_{nl}^{lift}=\chi_{nl}$ i.e., the class of each lifted vector remains the same, and $p_{nl}^{lift}$ is provided by \eqref{softmax_lift}. The derivative of the cross-entropy loss function $L_N^{lift}$ with respect to an element of the weight matrix $W^{lift}$ is given by 
\begin{equation}\label{loss_gradient_lift}
\frac{\partial L_N^{lift}}{\partial w_{ij}^{lift}} = - \frac{1}{N} \sum_{n=1}^N [\chi_{ni}^{lift} - p_{ni}^{lift}] v_{nj}^{lift}, \;\; \text{for} \;\; i=1,\ldots,m \;\; \text{and} \;\; j=1,\dots,m_{lift}.
\end{equation}

\begin{algorithm}[H] 
	\caption{Lifting-enhanced classification algorithm}
	\begin{algorithmic}[1]
		\Require Number of classes $m,$ data $V_N$ (vectors of dimension $m$) where  $N = m \times N_{class}$ ($N_{class}$ orthogonal vectors from each of the $m$ classes perturbed with $\mathcal{N}(0,\sigma^2)$ noise, $\sigma^2$ is user-prescribed), number of training data $N_{train}$ and test data $N_{test}$ with $N_{train} + N_{test} = N,$ class labels $\chi$ for the vectors in $V_N,$ lifting dimension $m_{lift}$ ($m_{lift} > m$),  mini-batch size $N_b,$ initial learning rate for Adam optimizer $\eta,$ number of training epochs $N_{epochs}$
		\State \textbf{Lifting:}
		\State Create the $m_{lift}$-dimensional lifted data $V_N^{lift}$ using \eqref{lift}
                 \State \textbf{Training:}
                 \State Initialization of the $m \times m_{lift}$ weight matrix $W^{lift}$
		\For{each training epoch $ n = 1, 2, \dots, N_{epochs} $}
		\State Pick a mini-batch from the training data $V_{N_{train}}^{lift}$
		\State Assign class probabilities to the mini-batch using a softmax classifier
		\State Use the Adam optimizer with cross-entropy loss function to update the weight matrix $W^{lift}$
		\State Repeat until the epoch is completed
        		\EndFor
	\end{algorithmic}
	\label{algorithm_lift}
\end{algorithm}

{\bf Chaos-enhanced model.} While lifting to higher dimensions can improve classification accuracy, even more significant improvement can be achieved if we further employ chaos. Specifically, each lifted vector, which is a point in an $m_{lift}$-dimensional space, can be identified as the initial state of a chaotic dynamical system and evolved for a prescribed temporal interval. The obvious question is why should such a process aid the classification accuracy? Intuitively, a chaotic dynamical system increases the distance of states that are initially close. In our case, the vectors resulting from the lifting process already contain randomness in the $m_{lift}-m$ dimensions. However, a controlled evolution under a chaotic dynamical system can pull those vectors apart, in essence homogenizing the randomness in all $m$ dimensions. This results in a more efficient use of the high dimensionality of the lifting space (this statement will become more concrete quantitatively in Section \ref{results}).

For our numerical experiments we have employed the Lorenz 96 model \cite{lorenz1996predictability}. The model is defined as
\begin{gather}
\frac{dx_i}{dt}= (x_{i+1}-x_{i-2})x_{i-1}-x_i + F, \;\; \text{for} \;\; i=1,\ldots,K, \label{lorenz1} \\ 
\text{with} \;\; x_{-1}=x_{K-1}, \;\; x_{0}=x_K, \;\; x_{K+1}=x_1, \;\; K \geq 4, \label{lorenz2}
\end{gather} 
and $F$ is a forcing term. Based on the literature, we have chosen the value $F=8$ for all the numerical experiments, although a more detailed study of the effect of $F$ on the results will be interesting \cite{karimi2010extensive}. Also, because of the lower bound on $K$ from \eqref{lorenz1}, the 3 periodicity conditions in \eqref{lorenz2} and the lifting construction, in our numerical experiments we consider lifting spaces of dimension $m_{lift} \geq m+6.$

An important question to ask is about the length of the temporal interval $T$ for which one should evolve the chaotic dynamical system so that it is beneficial for the classification accuracy \cite{vallejo2017predictability}. Based on dynamical system theory, a Lyapunov time is the time it takes for the deviation between two initial conditions to grow by a factor of $e.$ Thus, a few Lyapunov times is a good indicator of a chaotic system's predictability. Based on Lyapunov time estimates for the Lorenz 96 model (around 0.6 units of time for the regimes studied in the literature \cite{karimi2010extensive,brajard2020combining}), we chose $T=2$ units of time for our numerical experiments (except where noted, see also the discussion in Section \ref{selection} about a process for deciding the optimal value of $T$).

We denote the $n$-th lifted and chaos-evolved data vector as $v_{n}^{lift,chaos}.$ As a result, the logits $z_{ni}^{lift,chaos} = \sum_{j=1}^{m_{lift}} W_{ij}^{lift}v_{nj}^{lift,chaos}$ for $i=1,\ldots,m,$ and softmax outputs the $m$ probabilities 
\begin{equation}\label{softmax_lift_chaos}
p_{ni}^{lift,chaos} = \frac{e^{z_{ni}^{lift,chaos}}}{\sum_{i=1}^m e^{z_{ni}^{lift,chaos}}}, \;\; \text{for} \;\; i=1,\ldots,m,
\end{equation}
that the vector $v_n^{lift,chaos}$ belongs to each of the $m$ classes. The loss function is given by 
\begin{equation}\label{loss_lift_chaos}
L_N^{lift,chaos} = - \frac{1}{N} \sum_{n=1}^N \sum_{l=1}^m \chi_{nl}^{lift,chaos} \log p_{nl}^{lift,chaos},
\end{equation}
where $\chi_{nl}^{lift,chaos}=\chi_{nl}$ i.e., the class of each lifted and chaos-evolved vector remains the same, and $p_{nl}^{lift,chaos}$ is provided by \eqref{softmax_lift_chaos}. The derivative of the cross-entropy loss function $L_N^{lift,chaos}$ with respect to an element of the weight matrix $W^{lift,chaos}$ is given by 
\begin{equation}\label{loss_gradient_lift_chaos}
\frac{\partial L_N^{lift,chaos}}{\partial w_{ij}^{lift,chaos}} = - \frac{1}{N} \sum_{n=1}^N [\chi_{ni}^{lift,chaos} - p_{ni}^{lift,chaos}] v_{nj}^{lift,chaos},
\end{equation}
for $ i=1,\ldots,m $ and $ j=1,\dots,m_{lift}.$

\begin{algorithm}[H] 
	\caption{Lifting- and chaos-enhanced classification algorithm}
	\begin{algorithmic}[1]
		\Require Number of classes $m,$ data $V_N$ (vectors of dimension $m$) where  $N = m \times N_{class}$ ($N_{class}$ orthogonal vectors from each of the $m$ classes perturbed with $\mathcal{N}(0,\sigma^2)$ noise, $\sigma^2$ is user-prescribed), number of training data $N_{train}$ and test data $N_{test}$ with $N_{train} + N_{test} = N,$ class labels $\chi$ for the vectors in $V_N,$ lifting dimension $m_{lift}$ ($m_{lift} \geq 7),$ dimension of Lorenz 96 model $K$ ($K = m_{lift}-3$), forcing magnitude $F,$   temporal interval for chaotic evolution $T,$ timestep for chaotic evolution $\eta_{chaos},$ mini-batch size $N_b,$ initial learning rate for Adam optimizer $\eta,$ number of training epochs $N_{epochs}$
		\State \textbf{Lifting:}
		\State Create the $m_{lift}$-dimensional lifted data $V_N^{lift}$ using \eqref{lift}
		\State \textbf{Chaotic evolution:}
		\State Identify each datapoint in $V_N^{lift}$ as an initial condition for Lorenz 96 and evolve for $T$ units of time with timestep $\eta_{chaos}$ using       \eqref{lorenz1}-\eqref{lorenz2} to create the data $V_N^{lift,chaos}$
                 \State \textbf{Training:}
                 \State Initialization of the $m \times m_{lift}$ weight matrix $W^{lift,chaos}$
		\For{each training epoch $ n = 1, 2, \dots, N_{epochs} $}
		\State Pick a mini-batch from the training data $V_{N_{train}}^{lift,chaos}$
		\State Assign class probabilities to the mini-batch using a softmax classifier
		\State Use the Adam optimizer with cross-entropy loss function to update the weight matrix $W^{lift,chaos}$
		\State Repeat until the epoch is completed
        		\EndFor
	\end{algorithmic}
	\label{algorithm_lift_chaos}
\end{algorithm}

\section{Results}\label{results}

\subsection{Numerical experiments setup}
We have conducted numerical experiments to compare the baseline model, the lifting-enhanced model, and the lifting- and chaos-enhanced model for moderate number of dimensions (classes) $m=2,\ldots,20.$ Because we want to have an equal number of samples from each class, the total number of samples varies for the different values of $m.$ We chose $N_{class}=20$ samples from each class, so the number of samples for each value of $m$ is equal to $20 \times m.$ The variance of the random perturbations of the orthogonal vectors was set to $\sigma^2=10^{-4}$ for all the experiments (see also Section \eqref{selection} where we examine the effect of $\sigma^2$ on the accuracy). Also, we split the samples equally between those used for training and those used for testing. For the Adam optimizer, we set the mini-batch size to $N_b=10$ and learning rate $\eta=10^{-3}.$ We have added to the loss function of the baseline model,  the lifting-enhanced model and the lifting- and chaos-enhanced model the square of the $L_2$ norm of the weight matrix elements multiplied by $10^{-3}$ as a regularizer to the cross-entropy loss. The Lorenz 96 model was evolved with the standard 4th order Runge-Kutta scheme and $\eta_{chaos}=10^{-2}.$ All the calculations were performed in double precision.
%

Since we do not have yet a good way to decide on the optimal value of the lifting dimension, we conducted experiments where for each value of $m,$ we searched for the optimal lifting dimension $m_{lift}$ between the values $m+6$ and 50. The optimal value of $m_{lift}$ was decided based on an accuracy metric evaluated on the \textit{test} data. We chose the value of $m_{lift}$ that resulted in the highest testing accuracy during training. In our experiments, the optimal value of $m_{lift}$ fluctuated due to the stochasticity of the algorithm (random perturbations of the orthogonal vectors, randomness of the values of the components in the lifted dimensions and randomness of the Adam optimizer). A more detailed study of the optimal $m_{lift}$ will be presented elsewhere. 

There are different metrics for the accuracy of the predicted class probabilities. The \textit{proportion} accuracy metric is given by
\begin{equation}\label{proportion_accuracy_metric}
accuracy_{proportion} = \frac{1}{N} \sum_{n=1}^N \delta_{i_{max}i_n}, 
\end{equation}
where $i_{max}=\arg\max(p_{n1},p_{n2},...,p_{nm})$ is the predicted class for the $n$-th sample and $i_n$ is the actual class. For the proportion accuracy metric to be 1, it is only required that the predicted class probability for the correct class is larger than the predicted class probabilities for the rest of the classes. This metric is not stringent enough to distinguish between the various models. Instead, we have opted for the following accuracy metric
\begin{equation}\label{alignment_accuracy_metric}
accuracy_{alignment} = \frac{1}{N} \sum_{n=1}^N \chi_{n}\cdot p_{n} =  \frac{1}{N} \sum_{n=1}^N \sum_{l=1}^m \chi_{nl}p_{nl} = \frac{1}{N} \sum_{n=1}^N p_{ni_n},
\end{equation}
where $\chi_{nl} = \delta_{l i_n}$ (recall that $i_n$ is the correct class for the $n$-th sample) and $p_{nl}$ is the probability for class $l$ for the $n$-th sample predicted by one of the 3 models we compared (we note that $N$ in \eqref{alignment_accuracy_metric} stands for $N_{train}$ or $N_{test}$ depending on whether we measure training accuracy or testing accuracy). We see from \eqref{alignment_accuracy_metric} that to achieve 100\% accuracy, the predicted probability from the model for the correct label for each sample must be 1. Another way to interpret the metric in \eqref{alignment_accuracy_metric} is as confidence of the predicted class labels by the model, since it measures the \textit{alignment} between the predicted class probabilities vector and the correct class vector. Unless otherwise stated, the accuracy used in the figures is the alignment accuracy metric (although we utilize the proportion metric too in Section \ref{robustness} to discuss robustness).




\subsection{Observations from results}\label{observations} 
\begin{figure}[H]
\centering
    \begin{tabular}{cc}
  \includegraphics[width=0.40\textwidth]{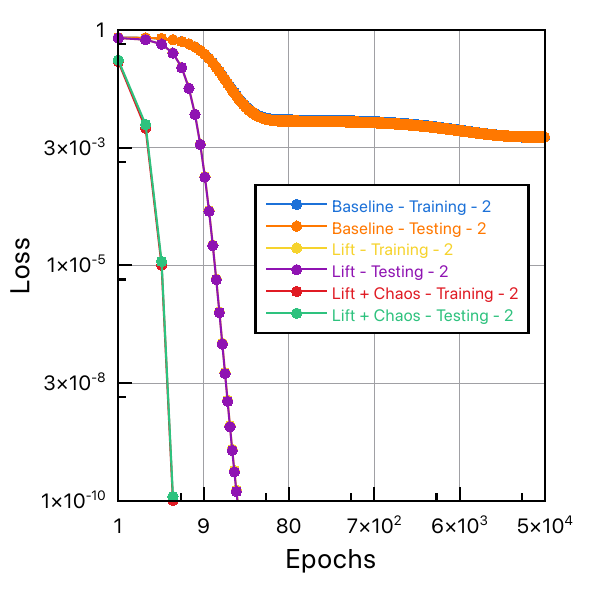} &
  \includegraphics[width=0.40\textwidth]{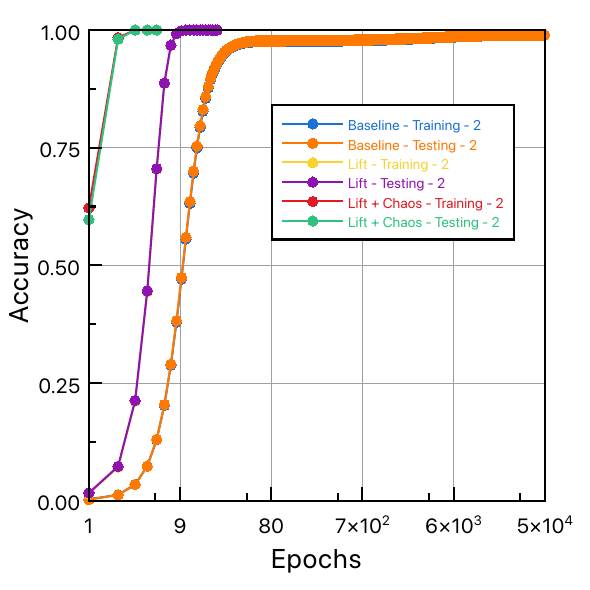}\\
   (a)  & (b)  \\
\end{tabular}
  \caption{$m=2.$ Comparison of the baseline model, the lifting-enhanced model, and the lifting- and chaos-enhanced model (with the optimal lifting dimension). (a) Evolution of loss (cross-entropy) with epochs. (b) Evolution of accuracy with epochs.}
  \label{fig:test_2}
\end{figure}

\begin{figure}[H]
\centering
    \begin{tabular}{cc}
  \includegraphics[width=0.40\textwidth]{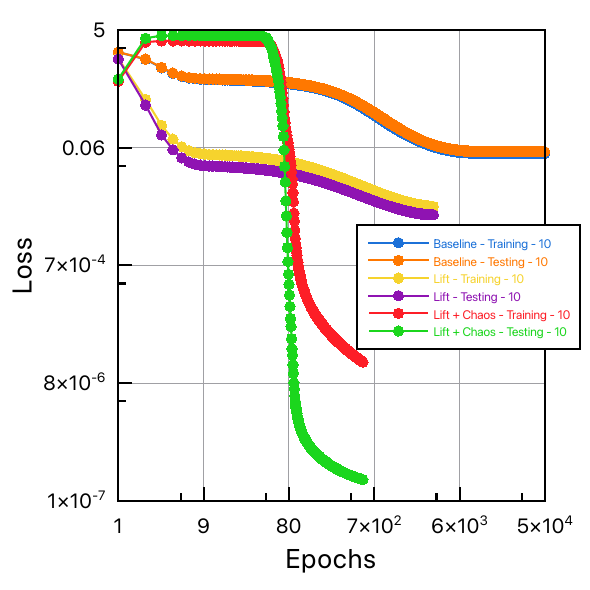} &
  \includegraphics[width=0.40\textwidth]{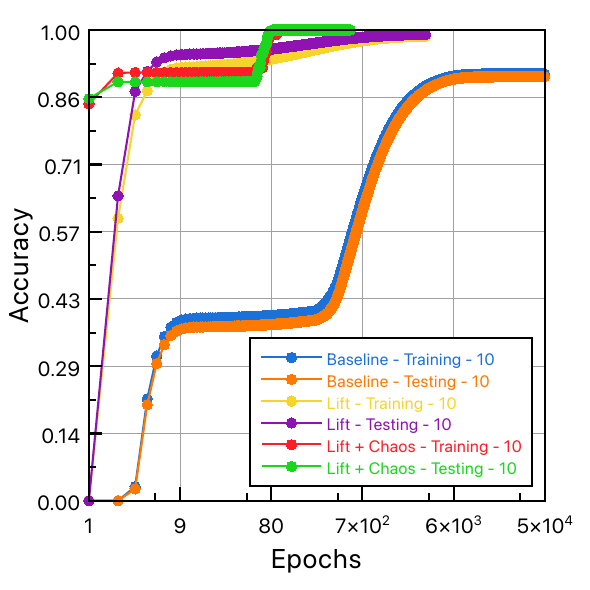}\\
   (a)  & (b)  \\
\end{tabular}
  \caption{$m=10.$ Comparison of the baseline model, the lifting-enhanced model, and the lifting- and chaos-enhanced model (with the optimal lifting dimension). (a) Evolution of loss (cross-entropy) with epochs. (b) Evolution of accuracy with epochs.}
  \label{fig:test_10}
\end{figure}

\begin{figure}[H]
\centering
    \begin{tabular}{cc}
  \includegraphics[width=0.40\textwidth]{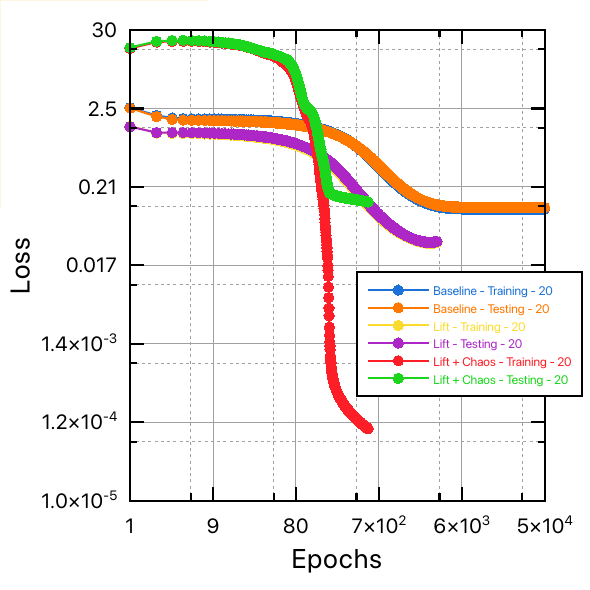} &
  \includegraphics[width=0.40\textwidth]{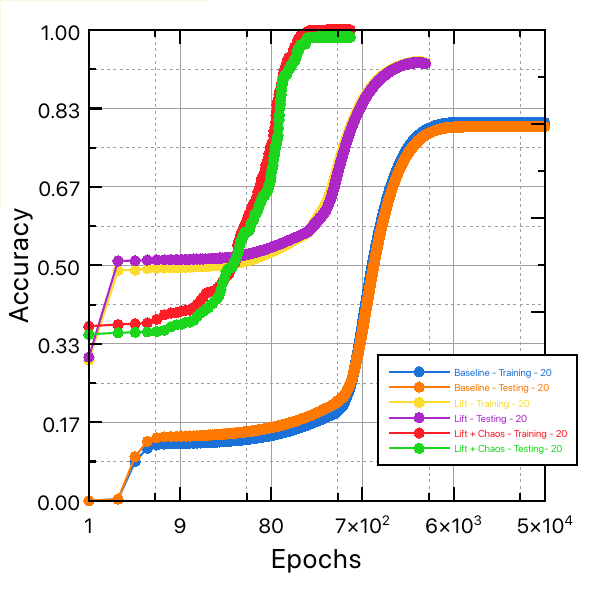}\\
   (a)  & (b)  \\
\end{tabular}
  \caption{$m=20.$ Comparison of the baseline model, the lifting-enhanced model, and the lifting- and chaos-enhanced model (with the optimal lifting dimension). (a) Evolution of loss (cross-entropy) with epochs. (b) Evolution of accuracy with epochs.}
  \label{fig:test_20}
\end{figure}

Figures \ref{fig:test_2}-\ref{fig:test_20} show the evolution of the loss (cross-entropy) and the accuracy with epochs for the training and testing for the 3 models when $m=2,10,20.$ Recall that we chose the optimal value of $m_{lift}$ based on the testing accuracy. The results lead to various observations about the performance of the 3 models.

First, while the lifting alone can improve performance over the baseline model, there is a dramatic acceleration of the decrease rate of the loss function and an equally dramatic acceleration of the convergence of the predicted accuracy to 1 (100\%) for the lifting- and chaos-enhanced model. This model can converge to perfect accuracy within a few epochs, while the corresponding loss is decreased to very small values. In fact, we had to put a lower threshold for the loss at $10^{-10}$ and terminate the algorithm when it was crossed, because the Adam optimizer sometime became unstable when the next epoch decreased the loss value from $10^{-10}$ to a value close to machine precision ($10^{-16}$) for our double precision calculations.

Second, the lifting- and chaos-enhanced model can reach perfect accuracy within a few hundred epochs (we allowed not more than 500 during our experiments), while the lifting-enhanced model, with the exception of $m=2,$ requires an order of magnitude more epochs to reach similar accuracy (if at all). For $m=2,$ the baseline model needs 20000 epochs to reach similar accuracy, while for $m=10$ and $m=20$ it is not able to do so.

Third, while the loss value decreases monotonically for the baseline and lifting-enhanced models for all values of $m,$ for the lifting- and chaos-enhanced model for $m=10$ and $m=20,$ the training and test loss values initially go through a phase of increase before they start decreasing. It is not accidental that during the same phase, the accuracy of the lifting- and chaos-enhanced model is lower than that of the lifting-enhanced model. This behavior is intriguing and needs to be investigated more, since the lifting- and chaos-enhanced model behaves in a way similar to some evolutionary models which perform excursions in the solution landscape before honing on an accurate one (called the steppingstone principle \cite{lehman2011abandoning}). Of course, in the present model, there is no exploration during the optimization phase, but it is possible that the chaotic evolution which precedes the optimization step creates an appropriate warping of the data that favors such kind of behavior. There is also the possibility that the initial increase in the training and test loss values is due to the presence of class outliers which confuse the classifier as we explain more in our next observation.


%
%
%
%

%

Fourth, we observe in Fig. \ref{fig:test_20} that the test loss value for the lifiting- and chaos-enhanced model is significantly larger than the training loss value, yet the testing accuracy does not suffer significantly (it plateaus at 98.5\%). This is likely due to an outlier which results in rather small correct class probability prediction and gets penalized logarithmically for the loss function but only linearly for the accuracy metric. The reason for the outlier could be tied to the temporal interval $T$ of chaotic evolution being too long for some samples so their class label may be mistakenly identified if they venture into the region corresponding to a different class (see also the results below for $T=1.5$ instead of $T=2$ and where this phenomenon does not occur and also the discussion in Section \ref{reason}). We have included this experiment on purpose to emphasize that these results do not involve any hyperparameter optimization and this is something that we will comment on briefly now and address with a more elaborate study appearing in future work. Specifically, in all the experiments presented in Figs. \ref{fig:test_2}-\ref{fig:test_20}, we have fixed the temporal interval of the chaotic evolution to $T=2$ and the learning rate of the Adam optimizer is $\eta=10^{-3}.$ Both choices were based on prior experience but they were not optimized in the usual manner. To show that the results of the lifting- and chaos-enhanced model can improve even more, we will start by examining sensitivity of the accuracy to the lifting dimension $m_{lift}$ for different values of $T$ and $\eta.$

\begin{figure}[H]
\centering
  \includegraphics[width=0.40\textwidth]{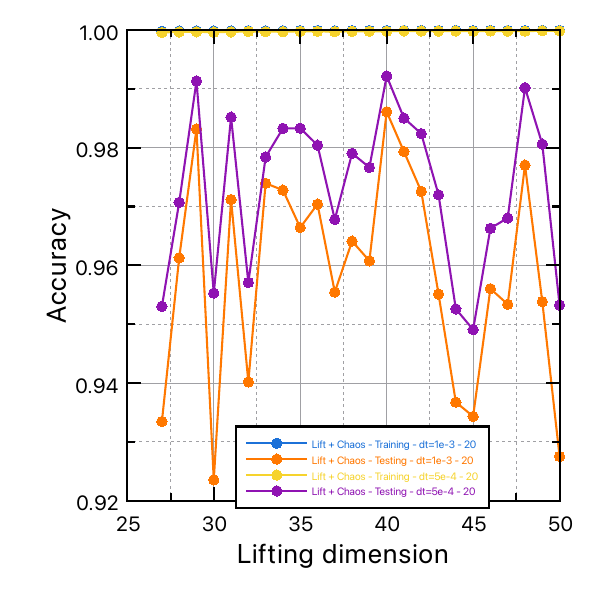} 
  \caption{$m=20.$ Evolution of training and testing accuracy with the lifting dimension for the lifting- and chaos-enhanced model for two different values of the Adam optimizer learning rate $\eta.$}
  \label{fig:test_accuracy_lifting_dimension_variable_eta_20}
\end{figure}

Figure \ref{fig:test_accuracy_lifting_dimension_variable_eta_20} shows the evolution of training and testing accuracy with the lifting dimension for the lifting- and chaos-enhanced model for two different values of the Adam optimizer learning rate $\eta,$ namely $\eta=10^{-3}$ and $\eta=5\times10^{-4}.$ We keep the maximum number of training epochs equal to 500 in both cases. We observe that while the training accuracy is equally good for the two learning rates, the testing accuracy is higher when $\eta=5\times10^{-4}.$ 

\begin{figure}[H]
\centering
\begin{tabular}{cc}
  \includegraphics[width=0.40\textwidth]{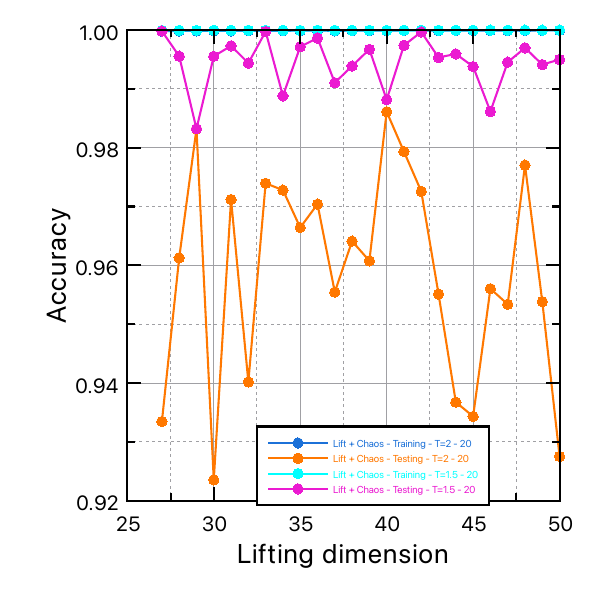} &
  \includegraphics[width=0.40\textwidth]{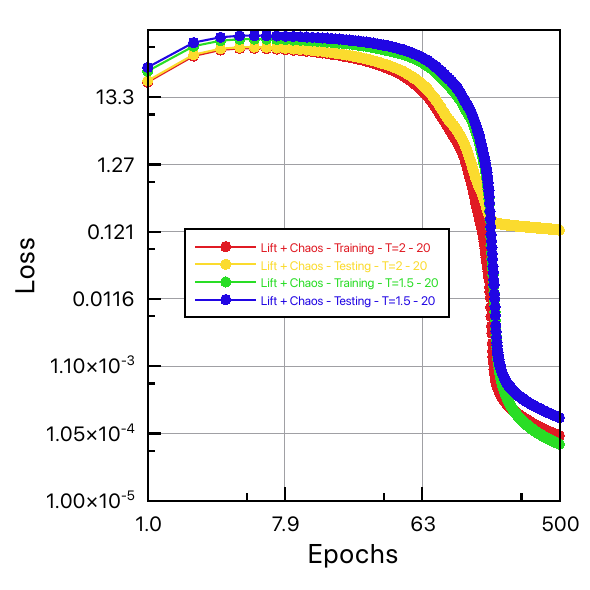} \\
  (a)  & (b)  \\
\end{tabular}
  \caption{$m=20.$ (a) Evolution of training and testing accuracy with the lifting dimension for the lifting- and chaos-enhanced model for two different values of the chaotic evolution temporal interval $T.$ (b) Evolution of training and testing accuracy with epochs for the optimal lifting dimension of the lifting- and chaos-enhanced model for two different values of the chaotic evolution temporal interval $T.$}
  \label{fig:test_accuracy_lifting_dimension_variable_interval_20}
\end{figure}

We continue with the examination of the effect of the chaotic evolution temporal interval value $T$ on the accuracy for different lifting dimensions. Fig. \ref{fig:test_accuracy_lifting_dimension_variable_interval_20}(a) shows the evolution of the training and testing accuracy with lifting dimension for two values of the temporal interval, namely $T=1.5$ and $T=2.$ We observe that $T=1.5$ leads consistently to higher testing accuracy than $T=2,$ implying better generalization capability. Also, Fig. \ref{fig:test_accuracy_lifting_dimension_variable_interval_20}(b) shows that for $T=1.5,$ the testing loss tracks closely the training accuracy, in contrast to the results for $T=2.$

\subsection{Reason for improvement over the baseline model}\label{reason} The improvement in classification accuracy that is evident for the lifting-enhanced model but especially for the lifting- and chaos-enhanced model warrants an explanation. We provide here a first attempt at explaining this improvement by examining (i) the distribution of the data to be classified after the lifting and chaotic evolution, and (ii) the distribution of the elements of the weight matrix. 

\begin{figure}[H]
\centering
    \begin{tabular}{cc}
  \includegraphics[width=0.40\textwidth]{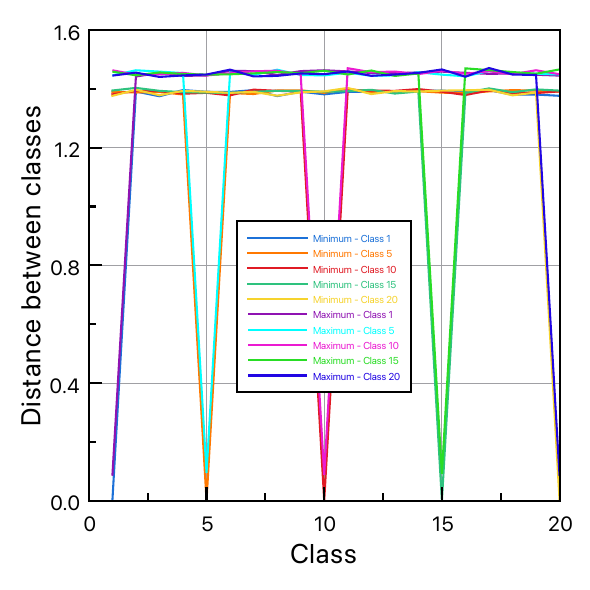} &
  \includegraphics[width=0.40\textwidth]{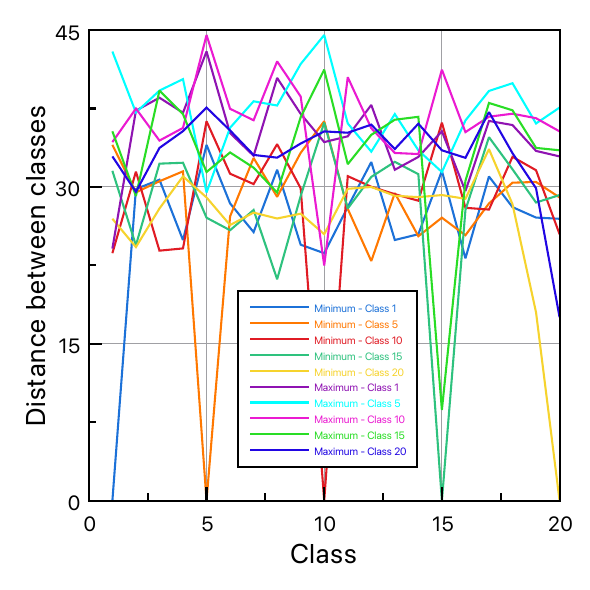}\\
   (a)  & (b)  \\
\end{tabular}
  \caption{$m=20.$ Minimum and maximum Euclidean distance between the clusters of points corresponding to the different classes. To avoid clutter we only plot the distances from other classes for the classes 1,5,10,15 and 20 (note the zero minimum distance for the classes 1,5,10,15 and 20 to themselves). (a) Baseline model. (b) Lifting- and chaos-enhanced model for $m_{lift}=26$ (the behavior is similar for higher lifting dimensions). Note the difference in scale in the two plots.}
  \label{fig:cluster_min_max_20}
\end{figure}

We begin with a comparison of the distribution of the data to be classified before and after the lifting and chaotic evolution. We look at the distribution of all the available data points i.e., the training and testing data because this will help explain not just the improved training accuracy but also the testing one. Fig. \ref{fig:cluster_min_max_20} shows the minimum and maximum Euclidean distance between the clusters of points corresponding to the different classes for the case of $m=20$ (to avoid clutter we only plot the distances from other classes for the classes 1,5,10,15 and 20, the behavior for the other classes is similar). From Fig. \ref{fig:cluster_min_max_20}(a), we see that before the lifting and chaotic evolution, the minimum and maximum distance is close to $\sqrt{2}$ which is to be expected because the data to be classified are perturbations of canonical basis vectors in $\mathbb{R}^m.$ On the contrary, from Fig. \ref{fig:cluster_min_max_20}(b), we see that the lifting and chaotic evolution increase significantly the minimum and maximum distance between the clusters of points corresponding to the different classes, with these distances fluctuating between 20 and 45. The increase in the minimum and maximum distances suggests that the clusters corresponding to the different classes are easier to distinguish and, subsequently, more amenable to unambiguous classification (see also the discussion in Section \ref{selection} about the effect of the perturbation variance $\sigma^2$ on the distance between clusters).

\begin{figure}[H]
\centering
  \includegraphics[width=0.40\textwidth]{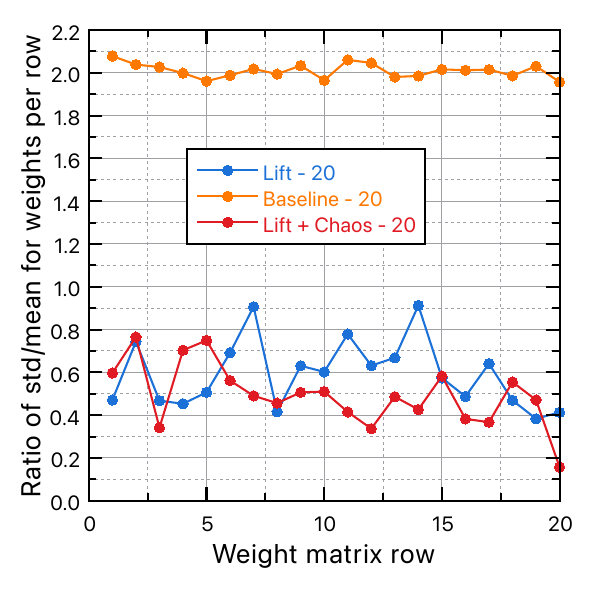} 
  \caption{$m=20.$ Ratio of the standard deviation over the mean of the absolute value of the components of each row of the weight matrix for the 3 models. Note that each row of the weight matrix corresponds to a class.}
  \label{fig:spread_20}
\end{figure}

We continue with an examination of the distribution of the weight matrix components for the baseline model, the lifting-enhanced and the lifting- and chaos-enhanced model.  For the baseline model, each row of the weight matrix is a vector in $m$-dimensional space and each row corresponds to one class. For the lifting-enhanced and the lifting- and chaos-enhanced models, each row of the weight matrix is a vector in $m_{lift}$-dimensional space and each row corresponds to one class. In Fig. \ref{fig:spread_20} we plot the ratio of the standard deviation over the mean of the absolute value of the weight matrix row components for each row (class) for $m=20.$ This metric shows how spread are the values of the weights in each row of the weight matrix. From Fig. \ref{fig:spread_20} we see that the ratio is much lower for the lifting-enhanced model and the lifting- and chaos-enhanced model than for the baseline model. This means that the components across each row of the weight matrix are much more similar for these two models than for the baseline model. Also, the ratio values are within a small range across rows (classes). A plausible explanation for this phenomenon is that the two models which operate in the $m_{lift}$-dimensional space converge through training to rows of the weight matrix (1 per class) which reside close to the corners of $m_{lift}$-dimensional hypercubes. In this way, the components of each row may have different signs but similar absolute value. As a result, the rows of the weight matrix (seen as vectors) spread apart in the $m_{lift}$-dimensional space and facilitate classification since it is easier to be distinct from one another. 

\begin{figure}[H]
\centering
    \begin{tabular}{cc}
  \includegraphics[width=0.40\textwidth]{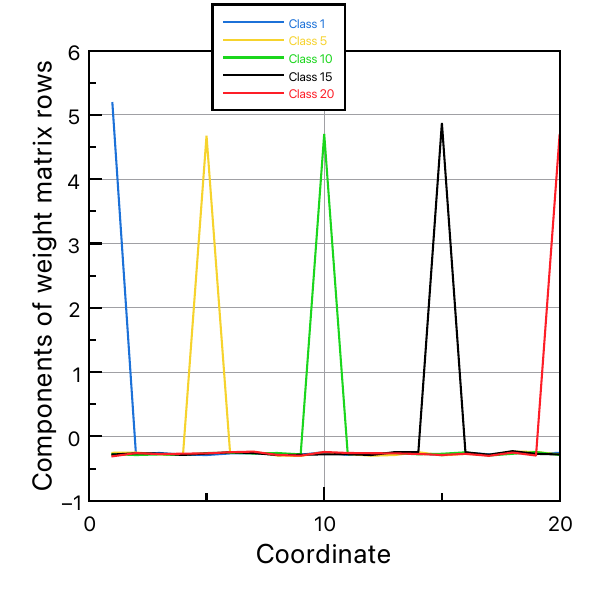} &
  \includegraphics[width=0.40\textwidth]{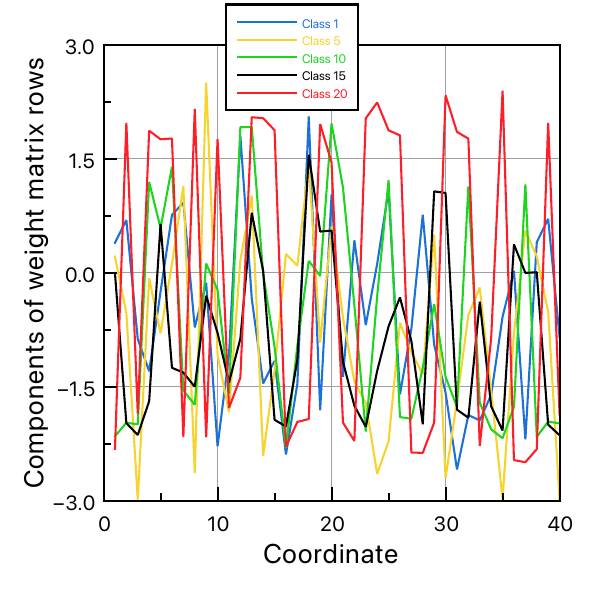}\\
   (a)  & (b)  \\
\end{tabular}
  \caption{$m=20.$ Components of rows of the weight matrix (each row corresponds to a class) for the baseline model and the lifting- and chaos-enhanced model. To avoid clutter we have plotted only the weights for rows (classes) 1,5,10,15 and 20. (a) Baseline model. (b) Lifting- and chaos-enhanced model. Note that the optimal $m_{lift}$ determined during training is 40 and that is why there are 40 elements (coordinates) of the weight matrix in each row.}
  \label{fig:weight_20}
\end{figure}

We plot in Fig. \ref{fig:weight_20} the value of the weights across a row of the weight matrix for different rows (recall that each row corresponds to 1 class). We plot this both for the baseline model and the lifting- and chaos-enhanced model. The difference between Fig. \ref{fig:weight_20}(a) and Fig. \ref{fig:weight_20}(b) is striking. For the baseline model, the weights corresponding to each class concentrate strongly in one element of the row (one coordinate in the $m$-dimensional space). This is understandable since the data we want to classify are perturbations of the canonical basis in $\mathbb{R}^m$ and the distribution of the weights values across a row reflects this structure. As expected, the weight matrix is very efficient in identifying data from this particular class, assuming they are not heavily perturbed. On the other hand, for the lifting- and chaos-enhanced model, the weights corresponding to each row (class) fluctuate across the row (the coordinates in the $m_{lift}$-dimensional space) and thus can handle data which are more strongly perturbed. This result suggests an increased robustness to noise in the data by the lifting- and chaos-enhanced model.

\subsection{Robustness}\label{robustness} 
\begin{figure}[H]
\centering
    \begin{tabular}{cc}
  \includegraphics[width=0.40\textwidth]{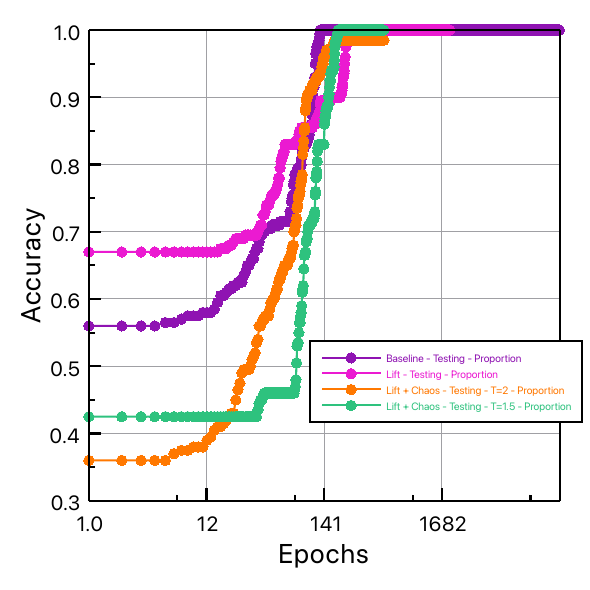} &
  \includegraphics[width=0.40\textwidth]{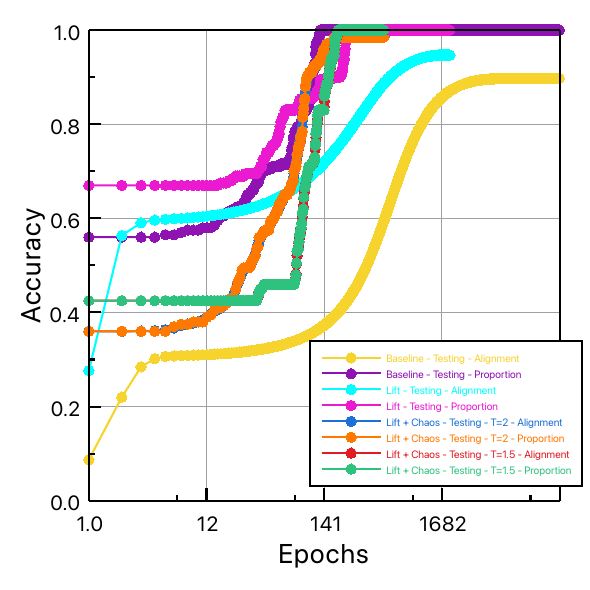}\\
   (a)  & (b)  \\
\end{tabular}
  \caption{$m=20.$ (a) Proportion accuracy metric evolution with epochs for the baseline, lifting-enhanced and lifting- and chaos-enhanced models (for two different chaotic evolution temporal interval values). (b) Proportion and alignment accuracy metric evolution with epochs for the baseline, lifting-enhanced and lifting- and chaos-enhanced models (for two different chaotic evolution temporal interval values).}
  \label{fig:robustness_20}
\end{figure}

We can provide a first insight into the increased robustness of the lifting- and chaos-enhanced model compared to the baseline and lifting-enhanced models by employing the proportion accuracy metric in \eqref{proportion_accuracy_metric} in addition to the alignment accuracy metric in \eqref{alignment_accuracy_metric}. Fig. \ref{fig:robustness_20}(a) shows the evolution of the testing proportion metric for the baseline, lifting-enhanced and lifting- and chaos-enhanced models (for two different chaotic evolution temporal interval values). The proportion metric behaves similarly for all the models in terms of limiting value and rate of convergence. 

However, the picture changes drastically when we include the alignment metric (as shown in Fig. \ref{fig:robustness_20}(b)). Specifically, while for the lifting- and chaos-enhanced model, the alignment metric evolution tracks the proportion metric evolution, for the baseline and lifting-enhanced models the alignment metric evolution lags significantly behind the proportion metric evolution. In addition, the plateau reached by the alignment metric for the baseline and lifting-enhanced models is significantly below 1 (100\%). This means that even though the baseline and lifting-enhanced models can predict 100\% of the correct labels for the test data (proportion metric equals 1), the corresponding confidence in the predictions is lacking. Moreover, if one stops training the baseline and lifting-enhanced models when the proportion metric reaches 1, which is after 130 epochs for the baseline model and 244 epochs for the lifting-enhanced model, respectively, the corresponding confidence in the predictions, given by the alignment metric is rather low, 36\% for the baseline model and 77\% for the lifting-enhanced model. 

The confidence in the predictions of the baseline model suffers due to the lack of alignment between the predicted class probability vectors with the correct label vectors. We attribute this fragility to the presence of noise in the data which makes the robust prediction of correct labels more difficult (recall the sharp concentration of the weight matrix rows on a single class as shown in Fig. \ref{fig:weight_20}(a)). The same argument applies to a lesser degree for the lifting-enhanced model which is more robust to noise, but still not as robust as the lifting- and chaos-enhanced model. A more detailed study to explain the agreement between the proportion and alignment metric for the lifting- and chaos-enhanced model will appear elsewhere. 

\subsection{Estimation of the optimal chaotic evolution interval}\label{selection}
The results presented in the previous sections show the promise of the lifting- and chaos-enhanced model in accelerating training and increasing accuracy for classification. However, we need to understand better the dependence of the proposed model on the length of the chaotic evolution interval $T.$ Our numerical experiments have hinted that the chaotic evolution interval should not be too short and it cannot be too long. If the interval is too short, the chaotic evolution acts effectively as extra perturbation of the data to be classified which, in turn, can degrade the classification accuracy. On the other hand, if the interval is too long, datapoints from one cluster (class) can wander off and blend with the datapoints from another cluster (class), again degrading the classification accuracy. In addition, the length of the interval must depend on the variance $\sigma^2$ of the noise with which we have perturbed the orthogonal vectors corresponding to the different classes. Indeed, if the noisy perturbation is too large, it can create blending of datapoints belonging to different classes. These observations mean that there is an intermediate regime of intervals $T$ for which, ideally, we would like to establish a process for choosing an optimal length from. 
\begin{figure}[H]
\centering
  \includegraphics[width=0.40\textwidth]{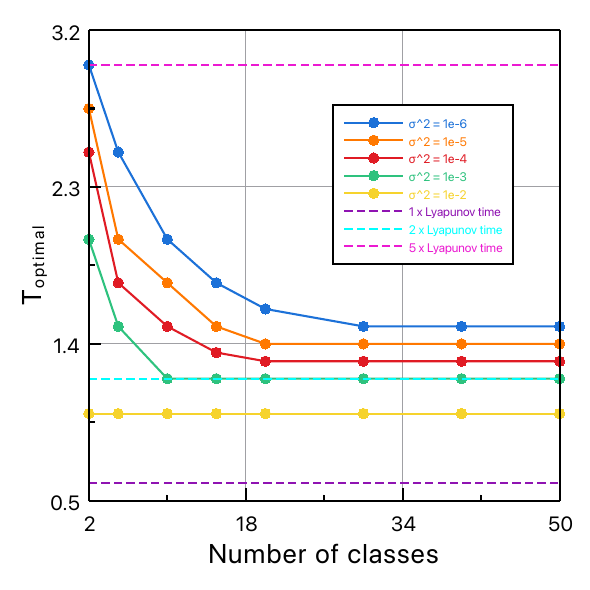} 
  \caption{Plot of the optimal chaotic evolution interval length as a function of the number of classes for different values of the perturbation variance $\sigma^2.$ Note that we have extended the number of classes up to 50. The plot includes also indicative multiples of the Lyapunov time for the Lorenz 96 model.}
  \label{fig:optimal}
\end{figure}

Fig. \ref{fig:optimal} shows the values of the optimal chaotic evolution interval $T_{optimal}$ as a function of the number of classes for different values of the perturbation variance $\sigma^2.$ Note that we have extended the number of classes up to 50 because they help with establishing the trends. The notion of optimality here is the largest allowed value of $T$ for which the results of the lifting- and chaos-enhanced model were superior than the baseline model both in terms of training efficiency and classification accuracy (including the total computational time needed to sweep a collection of lifting dimensions to pick the one for which we have the highest alignment training accuracy, see also discussion in Section \ref{complexity} below). Fig. \ref{fig:optimal} contains also markers of indicative multiples of the Lyapunov time $t_{L}=0.6$ for the Lorenz 96 model. We did not perform a thorough search for the $T_{optimal}$ value for each value of $\sigma^2$ and number of classes, we only examined cases that show the main trends. 

First, as expected the maximum allowed value of $T_{optimal}$ decreases both as a function of $\sigma^2$ and of the number of classes. Larger $\sigma^2$ means that the datapoints in a cluster will evolve for shorter time before they start blending with the datapoints from another cluster. In addition, since a higher number of classes $m$ means a higher starting dimension for the lifting and chaotic evolution through the Lorenz 96 dynamics, we expect that the optimal $T$ value will decrease as $m$ increases. This is because the Lorenz 96 model's complexity (the system's fractal dimension and number of positive Lyapunov exponents), increases with increased dimension even though it is a case of extensive chaos i.e., the intensity of chaos per dimension remains relatively stable as the system dimension increases \cite{karimi2010extensive}. Second, for 4 orders of magnitude of $\sigma^2,$
 namely from $10^{-6}$ to $10^{-3},$ the $T_{optimal}$ values fit in the interval $[2t_L,5t_L].$ When we increase the perturbation variance $\sigma^2$ up to $10^{-2}$ corresponding to 10\% noise, the $T_{optimal}$ drops to 1 for all the numbers of classes. Also, in this case, for the lifting- and chaos-enhanced model to perform better than the baseline model we have to increase the number of training epochs from 500 to 2000. If we want to increase the perturbation variance $\sigma^2$ even more, we have found that the baseline model or the lifting-enhanced only model behave better than the lifting- and chaos-enhanced model which is to be expected. The magnitude of $\sigma^2$ is large enough that there is no room for chaotic evolution before the datapoints from different classes start blending. Third, the value of $T_{optimal}$ appears to converge for the different values of $\sigma^2$ as we increase the number of classes. One interesting aspect of this convergence is that it is towards values close to $2t_L=1.2$ which signifies the weakening of predictability due to exponential separation of nearby initial conditions. We know that the datapoints belonging to a class are close by when seen as initial conditions for the Lorenz 96 model and thus a reasonable time length for their separation is expected to be a few multiples of the Lyapunov time. 
 
Despite the observation of trends in Fig. \ref{fig:optimal}, there does not appear to be an easily discernible criterion for the selection of $T_{optimal}.$ We discuss here a first attempt at establishing a selection process based on dynamical system theory considerations. Recall that the basic premise of the lifting- and chaos-enhanced approach is that it allows the clusters of points corresponding to different classes to move away from one another, as well as allow datapoints within a cluster to spread in a controllable fashion, thus making the classification task more efficient and accurate. However, as we have discussed, we expect a lower and upper bound on the range of values that we should look for $T_{optimal}$ in. This situation is reminiscent of intermediate asymptotics and we should decide what quantities are the most important in determining $T_{optimal}$ \cite{barenblatt2003scaling}.
 
 We have opted for measuring the ratio of two quantities that measure the distance between clusters and the spread of points within a cluster. Specifically, we monitor the ratio of the average distance from a cluster centroid over the centroid distance between clusters. In addition, because there are several clusters and also several lifting dimensions we perform an average of the ratio over those values.
  
 The centroid of the clusters after they have been evolved for $t$ units of time is defined as
\begin{equation}\label{centroid}
c_{class_i}(t)=[\frac{\sum_{j \in I_i}v^{lift}_{j1}(t)}{N_{class_i}},\ldots,\frac{\sum_{j \in I_i}v^{lift}_{jm_{lift}}(t)}{N_{class_i}}]^T,
\end{equation}
where $v^{lift}_j(t)=[v^{lift}_{j1}(t),\ldots,v^{lift}_{jm_{lift}}(t)]^T$ is the $m_{lift}$-dimensional vector of a point belonging to $class_i,$ the set $I_i$ contains the indices of all points in $class_i,$ and $N_{class_i}=|I_i|.$ We note that there is no confusion in \eqref{centroid} between the symbol used to denote the transpose of a vector and the length of the chaotic evolution interval $T.$
The average distance from the cluster centroid after $t$ units of time is defined as 
\begin{equation}\label{average}
average_{class_i}(t)=\frac{1}{N_{class_i}} \underset{l \in I_i}{\sum}\sqrt{\sum_{j=1}^{m_{lift}} [v^{lift}_{lj}(t)-c_{class_i,j}(t)]^2}.
\end{equation}

The centroid distance between clusters $i$ and $j$ after $t$ units of time is defined as 
\begin{equation}\label{centroid_distance}
d_{centroid}(class_i,class_j)(t)=\sqrt{\sum_{l=1}^{m_{lift}} [c_{class_i,l}(t)-c_{class_j,l}(t)]^2},
\end{equation}
where $i,j = 1,\ldots,m.$ 
We also define, for each class, the average cluster centroid distance after $t$ units of time given by
\begin{equation}\label{centroid_distance_average}
d_{centroid,i}(t)=\frac{1}{m-1}\sum_{\underset{i \neq j}{j=1}}^m d_{centroid}(class_i,class_j)(t),
\end{equation}
where $i=1,\ldots,m.$

As discussed above, we expect both the average distance from the cluster centroid and the distance between cluster centroids to increase with $t.$ We have decided to monitor the following ratio (and its first two derivatives)
\begin{equation}\label{ratio}
r(t) =\frac{1}{N_{lift}}  \sum_{j=1}^{N_{lift}} \frac{1}{m} \sum_{i=1}^{m} r_{i,j}(t)= \frac{1}{N_{lift}}  \sum_{j=1}^{N_{lift}} \frac{1}{m} \sum_{i=1}^{m}\frac{average_{class,i,j}(t)}{d_{centroid,i,j}(t)}, 
\end{equation}
where $average_{class,i,j}(t)$ is $average_{class,i}(t)$ for the $j$-th value of lifting dimension, $d_{centroid,i,j}(t)$ is $d_{centroid,i}(t)$ for the $j$-th value of lifting dimension, and $N_{lift}={m_{lift}}_{max}-{m_{lift}}_{min}+1$ is the total number of different cases of lifting dimension e.g., when we sweep the dimensions $(m+6,50)$ as we have indicated for many numerical experiments, $N_{lift}=50-(m+6)+1=45-m.$ The ratio $r(t)$ defined in \eqref{ratio} is an average of the ratios $r_{i,j}(t)$ over different classes and lifting dimensions. This choice is meant to act as a smoothing operation given the observed variation of the ratios $r_{i,j}(t)$ across classes and lifting dimensions. 

We expect the ratio $r(t)$ to start increasing after some lower bound of the chaotic evolution interval is crossed due to the fact that the chaotic evolution of the clusters will spread the datapoints far enough so that their distance from the cluster centroid will begin approaching the distance between cluster centroids. The first and second temporal derivatives of $r(t),$ denoted by $r'(t)$ and $r''(t),$ respectively, can help monitor the increase in $r(t)$ and will allow us to define a process for the estimation of the optimal chaotic evolution interval. 

We have chosen a one-sided finite difference approximation for the first two derivatives of $r(t).$ However, because the results can be noisy, we perform smoothing based only the nearest-neighbor values (techniques that produce even smoother derivative estimates will be explored in future work). Specifically, $r(t)$ is smoothed to $r_{smooth}(t)$ before used to compute $r'(t).$ Similarly, $r'(t)$ is smoothed to $r'_{smooth}(t)$ before used to compute $r''(t).$ Finally, we perform smoothing on $r''(t)$ to obtain $r''_{smooth}(t)$ before we use it. In our numerical experiments, we observed the following stages in the behavior of the smoothed versions of $r(t)$ and its first two derivatives. First, an initial period where $r'_{smooth}(t)$ fluctuates around 0. After that, $r'_{smooth}(t)$ starts increasing monotonically. Second, because of the monotonic increase of $r'_{smooth}(t),$ the second derivative $r''_{smooth}(t)$ is positive. In addition, due to the observed accelerating pace of increase of $r'_{smooth}(t),$ the value of $r''_{smooth}(t)$ is increasing. Third, as expected, there is a later time, when $r''_{smooth}(t)$ reaches a peak. Based on these observations, we have devised the following process to estimate $T_{optimal}.$ \\

\begin{figure}[H]
\centering
  \includegraphics[width=0.40\textwidth]{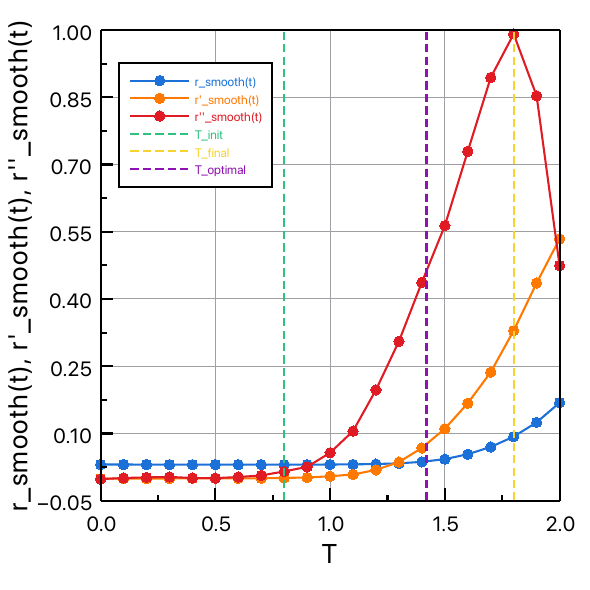} 
   \caption{$m=20.$ Plot of $r_{smooth}(t), r'_{smooth}(t), r''_{smooth}(t)$ as a function of $T$ for $\sigma^2=10^{-4}.$ The locations of the dashed vertical lines corresponding to $T_{init},$ $T_{final}$ and $T_{optimal}$ were determined using the selection process described in the text.}
  \label{fig:selection}
\end{figure}

{\bf Process of estimating $T_{optimal}$} 
\begin{itemize}
\item
Identify $T_{init}$ as the time that $r'_{smooth}(t)$ starts to be positive (after the initial period of fluctuations around 0).
\item
Identify $T_{final} \geq T_{init}$ as the time when $r''_{smooth}(t)$ reaches its first peak. 
\item
Collect the values of $r_{smooth}(t)$ for chaotic evolution interval values in $[T_{init},T_{final}],$ say $r_{smooth}(t_i)$ for $i=1,\dots,n_{opt}$ with $t_i \in [T_{init},T_{final}].$ Normalize $r_{smooth}(t_i)$ to be in $[0,1]$ i.e., define $$w_i = \frac{r_{smooth}(t_i)}{\sum_{i=1}^{n_{opt}}r_{smooth}(t_i)}$$ with  $i=1,\ldots,n_{opt}.$ Then, our optimal time estimate is given by $$T_{optimal}=\sum_{i=1}^{n_{opt}} w_i t_i .$$
\end{itemize}
Fig. \ref{fig:selection} shows the results of applying our optimal chaotic interval selection approach for $m=20$ and $\sigma^2=10^{-4}.$ We find $T_{init}=0.8,$ $T_{final}=1.8$ and $T_{optimal}=1.42.$ Given the number of factors that affect the evolution of $r(t),$ it is remarkable that our estimate of $T_{optimal}$ is rather close to the estimate 1.3 shown in Fig. \ref{fig:optimal} that was obtained through trial and error. Moreover, the results of the lifting- and chaos-enhanced model for $T=1.42$ are equally good with those for $T=1.3,$ providing a significant improvement on the results of the baseline model in terms of efficiency and accuracy, namely reducing the cross-entropy loss by several orders of magnitude and the number of training epochs by at least an order of magnitude.

\subsection{Complexity}\label{complexity} Finally, we comment on the computational complexity of the baseline model, the lifting-enhanced model and the lifting- and chaos-enhanced model. Table \ref{table_complexity} presents the computational times in seconds for the three models for $m=2,10,20$ (our calculations were run on an Apple M4 processor). We note that while for the baseline model we perform training for the dimension $m$ for 20000 epochs, for the lifting-enhanced and the lifting- and chaos-enhanced models we train for a collection of lifting dimensions, in the range $(m+6,50).$ The computational times reflect the \textit{total} time needed to train for all the allowed lifting dimensions. In addition, we have put an upper limit on the number of training epochs, 2000 epochs per lifting dimension value for the lifting-enhanced model and 500 for the lifting- and chaos-enhanced model. Upper limit means that the optimizer may not complete the allowed number of epochs. The reason this can happen is that the Adam optimizer was terminated if the loss value fell below $10^{-10}$ to avoid instability due to very small numbers. 

For $m=10,20,$ the lifting- and chaos-enhanced model requires more time to sweep across all the allowed lifting dimension cases and pick the optimal one than a single run of the baseline model. However, for the values of $m$ reported in Table \ref{table_complexity}, the baseline model accuracy does not improve even if we let it train for time comparable to the one we need for the training of the lifting- and chaos-enhanced model. In addition, the alignment testing accuracy of the lifting- and chaos-enhanced model with $T=2$ is  100\% for $m=2,10$ and 98.61\% for $m=20,$ while it is 100\% for $m=2,10,20$ when $T=1.5.$ The alignment testing accuracy of the baseline model is 99.94\% for $m=2,$ 95\% for $m=10,$ and 90\% for $m=20.$ Moreover, the robustness of the performance of the lifting- and chaos-enhanced model to the lifting dimension value, is a promising sign that one may not need to sweep a large number of lifting dimensions to find one which leads to high accuracy. With this in mind, the computational complexity of the lifting- and chaos-enhanced model becomes competitive to the baseline model while offering consistently superior accuracy. 

 
 \begin{table}[H]
\begin{tabular}{|l|*{4}{c|}}
\hline
\diagbox{$m$}{Model} & \makebox[7em]{Baseline} & \makebox[7em]{Lift} & \makebox[7em]{Lift + Chaos} \\
\hline\hline
2 & 0.28s  & 0.36s& 0.25s  \\
\hline
10 & 13s & 110s& 29s\\
\hline
20 & 151s& 570s& 170s  \\
\hline
\end{tabular}
\caption{Computing times for the 3 models (in seconds). The baseline model training runs for 20000 epochs, the lifting-enhanced model runs for up to 2000 epochs (per lifting dimension value) and the lifting- and chaos-enhanced for up to 500 epochs (per lifting dimension value). Note that for the lifting-enhanced and the lifting- and chaos-enhanced models, the computing time is the total time required to sweep through the lifting dimensions $(m+6,50)$ to determine the optimal one.}\label{table_complexity}
\end{table}


\section{Discussion and future work}\label{discussion}

We have presented an approach to enhance classification accuracy that uses a two-stage approach where the input vector is first lifted to a higher dimension and then is evolved for a fixed temporal interval through a chaotic system. We observe that the lifting to higher dimensions, and especially the chaotic evolution, can accelerate significantly the training process and in addition increase the achieved accuracy compared to a baseline model. The rationale behind the combination of lifting to higher dimensions and chaotic evolution is to explore the available space in higher dimensions to pull apart the clusters of points corresponding to the various classes, but to do so in a controlled manner. Specifically, we can consider the evolution by a chaotic system as taking the cluster of points in the higher dimensional space corresponding to a class and blowing it up so that it occupies a larger volume. This, in turn, aids the classifier by making it easier to distinguish each class from the others. Indeed, we have found that the lifting and chaotic evolution can lead to (i) moving apart of the clusters of points corresponding to the different classes, and (ii) much more uniform distribution of weight values within a row of the weight matrix and also across rows (classes) compared to the baseline model. While these observations provide some insight into the improved performance of the proposed approach, a more thorough analysis based on the dynamics of training should be undertaken. Specifically, we need to understand how the lifting and chaotic evolution affect the gradient of the loss function. 

Our results indicate that even though one can obtain dramatic increase in the classification performance without thorough hyperparameter optimization, the benefits to be reaped can increase even more through a careful analysis of the main elements of the proposed approach. These include the specific construction of the lifting to higher dimensions, the choice of the chaotic system to evolve the lifted vector as well as the temporal interval of the chaotic evolution (we have offered here a first attempt at a selection process for the optimal chaotic evolution interval). In addition, the use of more efficient stochastic optimizers than the standard Adam optimizer should be considered. Also, it will be interesting to explore connections with physical reservoir computing approaches e.g., \cite{fernando2003pattern,tanaka2019recent,marcucci2020theory}. 

Finally, the results presented here are only proof-of-concept. A much more substantial test will be to see whether the proposed approach can improve the classification accuracy for challenging datasets e.g., standard benchmarks from image processing \cite{deng2012mnist}.

\section*{Acknowledgements}
I would like to thank Shady Ahmed and Saad Qadeer for useful discussions and comments. This work is partially supported by the U.S. Department of Energy, Office of Science, Advanced Scientific Computing Research
program under the "LEADS: LEarning-Accelerated Domain Science" project (Project No. 85462). Pacific Northwest National Laboratory is a multi-program national laboratory operated for the U.S. Department of Energy by Battelle Memorial Institute under Contract No. DE-AC05-76RL01830.

\bibliographystyle{unsrt}
\bibliography{theory}

\end{document}